# Exploring the Potential of Latent Embeddings for Sea Ice Characterization using ICESat-2 Data

Daehyeon Han[1], Morteza Karimzadeh[1]

[1]Department of Geography, University of Colorado Boulder

*Abstract*— The Ice, Cloud, and Elevation Satellite-2 (ICESat-2) provides high-resolution measurements of sea ice height. Recent studies have developed machine learning methods on ICESat-2 data, primarily focusing on surface type classification. However, the heavy reliance on manually collected labels requires significant time and effort for supervised learning, as it involves cross-referencing track measurements with overlapping background optical imagery. Additionally, the coincidence of ICESat-2 tracks with background images is relatively rare due to the different overpass patterns and atmospheric conditions. To address these limitations, this study explores the potential of unsupervised autoencoder on unlabeled data to derive latent embeddings. We develop autoencoder models based on Long Short-Term Memory (LSTM) and Convolutional Neural Networks (CNN) to reconstruct topographic sequences from ICESat-2 and derive embeddings. We then apply Uniform Manifold Approximation and Projection (UMAP) to reduce dimensions and visualize the embeddings. Our results show that embeddings from autoencoders preserve the overall structure but generate relatively more compact clusters compared to the original ICESat-2 data, indicating the potential of embeddings to lessen the number of required labels samples.

## I. Introduction

Sea ice plays a crucial role in energy circulation and climate change beyond the polar regions [1, 2]. Remote sensing approaches have become essential for investigating sea ice due to the remote and harsh conditions in sea ice areas. Accurate measurement of sea ice enhances our understanding of heat exchange, ice deformation, and climate change. Ice, Cloud, and Elevation Satellite-2 (ICESat-2), launched in September 2018 by NASA, provides valuable surface elevation data through its high-precision green laser (532 nm) altimetry system [3]. Along-track elevation measurements of ICESat-2 have enabled various applications including ice surface classification, freeboard measuring, detection of melt ponds, and analysis of sea ice thermodynamics [4].

Recent studies have investigated machine learning applications for ICESat-2 sea ice data. The primary focus has been surface type classification [5], leveraging the high-resolution capabilities of ICESat-2. These approaches have demonstrated improvements in classifying open water, grey ice (thin ice), and sea ice. However, while achieving high accuracy, current methods heavily depend on manual sampling based on matched pairs with optical imagery, such as ICESat-2 and Sentinel-2 coincident observations. This process requires significant time and effort, as well as has limitation due to the scarcity of overpass coincidences in polar regions and weather condition. The requirement of manual sampling significantly limits the scalability of machine learning applications for ICESat-2.

Transfer learning, which involves pre-training models on another task, offers a promising solution to the small sample size limitation [6]. The key advantage of transfer learning lies in obtaining meaningful embeddings (i.e., representation) on another task for which more labeled data is available. Then, the representations can be transferred to another task with fine tuning, which enable achieving higher model performance with fewer labeled samples compared to training from scratch. Unsupervised learning takes a slightly different approach of using the same data source and deriving the learned embeddings without supervision on a specific task. This can be accomplished through the use of autoencoders or reconstruction algorithms. Through training on vast amounts of unlabeled data, this approach allows models to capture broader information patterns and improves performance when trained on target data [7].

Despite this potential, there has been a lack of studies on unsupervised learning with ICESat-2 data. [8] attempted to use an autoencoder with ICESat-2 data to denoise atmospheric profiles. To address the challenges of manual sampling, [9] employed unsupervised clustering with a Gaussian mixture model (GMM) to generate surface type labels for lead and non-lead. The total of 717,009 samples were generated by manually assigning 80 clusters from GMM. However, to the best of our knowledge, no previous work has attempted to derive latent embeddings directly from ICESat-2 data from unsupervised learning.

In this study, we investigated the effectiveness of unsupervised learning for ICESat-2 data to derive the learned embeddings. We first generated 100 m sequences of ATL07 tracks to capture topographic features, including surface height, photon rate, normalized background rate, and number of pulses. For embedding training, autoencoder models with encoder-decoder structures were designed using Long Short-Term Memory (LSTM) and 1-D Convolutional Neural Networks (CNN). These models are trained to reconstruct topographic sequences while learning compact representations in the bottleneck layer. The dimension of extracted embeddings was reduced using Uniform Manifold Approximation & Projection (UMAP) [10]. The comparison was conducted within the UMAP visualization space.

## II. Data and Methodology

### A. ICESat-2 ATL07

NASA provides sea ice height product of ICESat-2 Advanced Topographic Laser 07 (ATL07) product, which processes ATL03 photon height specifically for sea ice areas.

Table 1: The number of ICESat-2 ATL07 tracks and the corresponding generated topographic sequence with a 100 m length.

|  | Train set | Validation set | Test set | Total |
|---|---|---|---|---|
| ICESat-2 ATL07 tracks | 268 | 39 | 77 | 384 |
| # of topographic sequence | 11,954,622 | 1,129,695 | 2,938,119 | 16,022,722 |

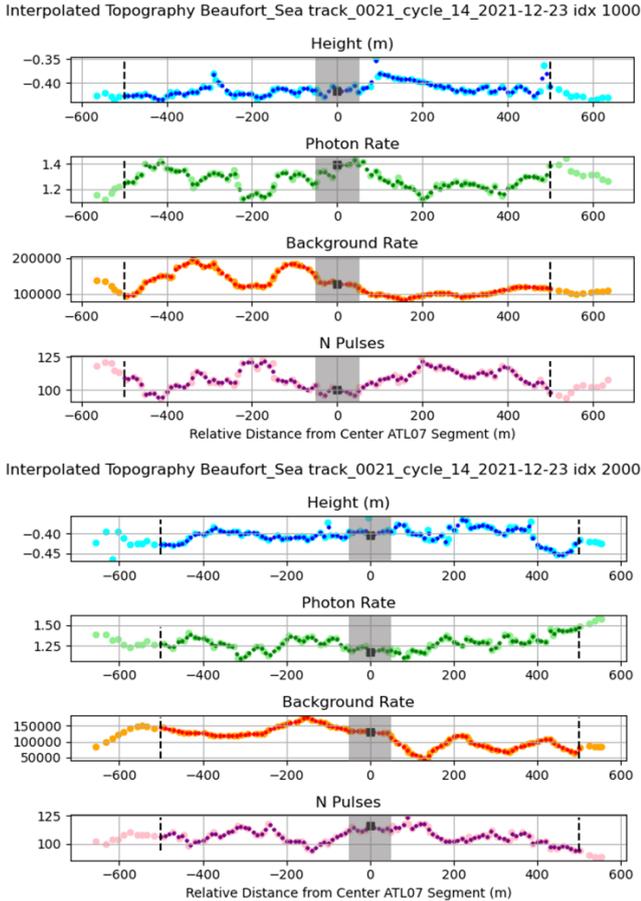

Fig. 1. Examples of interpolated topography data using a moving window approach. Light-colored points represent the original ATL07 values, while deep-colored points indicate the interpolated values with 10 m intervals. The vertical black dashed lines at -500 m and 500 m stand for the boundaries for trimming to generate 1-km topography for each ATL07 point. The gray shading between -50 m and 50 m represents the 100 m topographic sequence used for training and extracting embeddings in this study.

ATL07 aggregates approximately 150 photons to generate segments spanning 10 to 200 m [4]. The segment length varies based on atmospheric conditions and surface type. For our dataset, we collected 384 ATL07 tracks over the pan-Arctic area from December to February in 2018-2024. For spatiotemporal evaluation, the data was split into exclusive tracks with a ratio of 0.7, 0.1, and 0.2 for the Train, Validation, and Test sets, respectively (Table 1). Due to its high sensitivity to atmospheric conditions and sunlight, the normalized background rate per each ICESat-2 track was used, as highlighted in a key study on ICESat-2 surface classification using machine learning [5].

### B. Generating topography with super-segment

To extract embeddings from ATL07, we constructed surface topography by aggregating multiple ATL07 segments into super-segments (segment of segments). Given the variable intervals between ATL07 sequences, we implemented 10 m interval interpolation to ensure uniform spacing to count the distance in each sequence. This process used a moving window approach with forward and backward buffers from the center point. Super-segments containing any neighboring ATL07 distances exceeding 50 m were discarded to ensure data consistency and minimize uncertainty during interpolation. With 50 m of forward and backward buffers, each sequence consists of 11 points including the center point spanning 100 m (Fig. 1). Fig. 2 shows the overall distribution of each feature, demonstrating a similar pattern even with the exclusive splitting of non-overlapping ICESat-2 tracks.

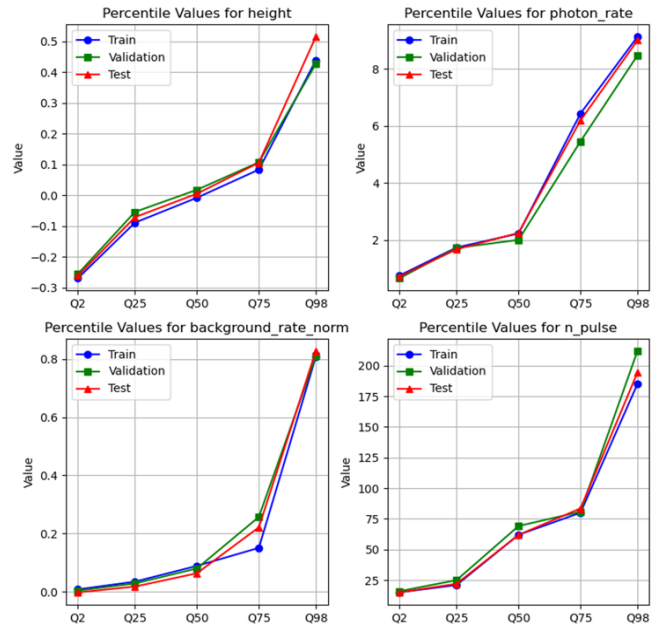

Fig 2. Distribution of input features for the Train, Validation, and Test sets.

### C. Autoencoder models

For embedding extraction, we designed two encoder-decoder autoencoder architectures. The first architecture employs LSTM, a representative sequence modeling approach, while the second utilizes 1-D CNN for capturing local features. The LSTM autoencoder consists of two LSTM layers in both encoder and decoder parts, with the encoder of 32 and 64 channels and the decoder containing 64 and 32 channels respectively. The CNN autoencoder maintains an identical structure but replaces LSTM layers with 1-D CNN layers. Between the encoder and decoder, a bottleneck layer with 16 channels exists for compact representations.

Non-linear activation functions were employed, with the hyperbolic tangent (tanh) function applied to the LSTM layers and the Rectified Linear Unit (ReLU) to the CNN layers. To prevent overfitting during reconstruction, a dropout rate of 0.2 was used for regularization for every LSTM and CNN layer. The output layer utilizes linear activation to reconstruct the input topography. For model optimization, we employed Mean Squared Error (MSE) as the loss function and trained using the Adaptive Momentum (Adam) optimizer [11]. The training process continued for a maximum of 1000 epochs with a batch size of 1024, implementing early stopping when validation loss showed no improvement for 20 consecutive epochs to prevent overfitting.

*D. Visualization of embeddings on UMAP space*

After completing the model training phase, we kept only the encoder and bottleneck components for embedding extraction. Then, samples were fed to the model to derive the learned embeddings. To visualize the embeddings in two dimensions, we used dimensionality reduction with UMAP, transforming from 16 to 2 dimensions with hyper parameters minimum distance=0.0001 and number of neighbors=50. We selected these hyper parameters after exploring multiple values with grid search approach. This reduction also facilitates the visualization of clustering results in a two-dimensional space.

*E. Experiment schemes*

Our experimental framework encompassed three schemes for visual evaluation with UMAP: the original ATL07 data, LSTM embedding, and CNN embedding. This comparison aims to evaluate how the optimized embedding with unsupervised learning can contribute to generating more compact and distinct clusters as visualized by UMAP results, potentially serving as a precursor for improved supervised learning with fewer samples. We evaluated the autoencoder reconstruction performance using both Root Mean Squared Error (RMSE) and $R^2$-score metrics. For the UMAP analyses, we utilized a subset of the training data (~100K samples) to improve computational efficiency and find the optimal parameters. We used an Nvidia A6000 GPU to train our autoencoder models. Training the LSTM model took about 2.5 hours, while the CNN model took around 20 minutes.

## III. RESULTS AND DISCUSSION

*A. Reconstruction performance of autoencoders*

Fig. 3 shows the reconstruction performance of autoencoder models for Test set. Overall, the reconstructed topography shows high agreement with the original sequence. LSTM achieved better performance than CNN in both RMSE and $R^2$-score for all four topographic features. CNN demonstrated relatively lower performance in height reconstruction, with an $R^2$ score of 0.8173, while the $R^2$-scores for other features were higher than 0.96. This suggests that the autoencoder models in this study have a strong ability to capture essential information in the embedding and accurately reconstruct the entire sequence from it.

*B. UMAP results and visualization*

Fig. 4 compares the visualization results of UMAP for a subset of Train set. UMAP distribution with ATL07 original input shows three major clusters: two larger clusters and one smaller cluster. UMAP results of both LSTM and CNN embeddings represent a similar pattern, also showing two larger clusters and one or two smaller clusters. However, UMAP results using LSTM and CNN embedding show more compact structures compared to the Original Input. For LSTM embedding, the two larger clusters are less clearly separated than in the Original Input. However, the LSTM clusters had L-shaped structures, which present more compact shape compared to the evenly scattered distribution of the Original Input over a wide area. On the other hand, the CNN embedding result in clearer separations between the major clusters, similar to the Original Input, while also having a more compact distribution.

Given that sea ice surface types are usually categorized into three major classes (water, thin ice, and sea ice) [5], we suspect that the major clusters observed in each scheme correspond to these classes. However, further investigation is required to confirm these findings and assess their practical implications. For more detailed classifications involving minor surface types, the more compact embedding distribution may lead to better clustering results. This suggests that embeddings generated by LSTM or CNN could potentially serve as useful inputs for clustering or classification tasks for downstream tasks.

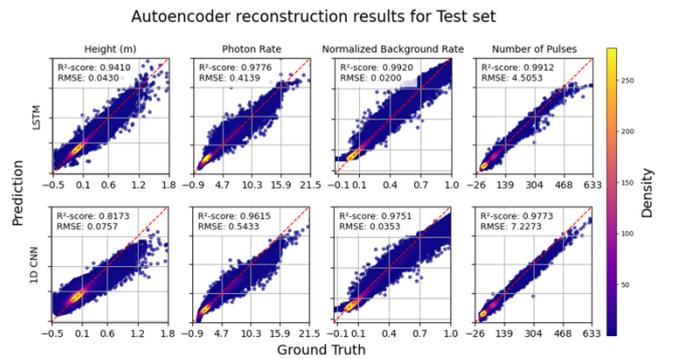

Fig. 3. Reconstruction results of topographic features using LSTM and CNN autoencoder models for the test set. The colors in the scatter plots and color bar indicate relative density of points in each subplot. Higher values (yellow) indicate areas where more scatter points overlap, representing higher density.

*C. Novelty, limitation, and future directions*

This study explored of the quality of embeddings generated by LSTM and CNN autoencoders compared with raw input (e.g., track observations) through UMAP analysis. There has been a lack of research leveraging unsupervised learning without labels for ICESat-2 data over sea ice regions. We expect that this study will draw attention to the potential of unsupervised learning to solve the challenges of manual sampling of ICESat-2 track data, which highly relies on the availability of background optical imagery.

Despite the improved compactness in UMAP results with embeddings, the absence of labeled data for UMAP visualization remains a key limitation of this study. This restricts the evaluation of how well the embeddings align with actual surface types or other meaningful classifications. Future evaluations incorporating labeled data will address this limitation, a direction that we are actively pursuing.

One potential reason for the high degree of similarity between the Original Input clusters and latent embedding clusters learned from LSTM or CNN can be traced in the ATL07 data. It might be that tracks over sea ice have generally highly-structured yet simple profiles, which is already captured in the 4-dimensional inputs (of height, photon rate, background rate, and number of pulses). Autoencoders are particularly good at extracting hidden patterns from complex and noisy data, which may not be the case for ICESat-2 tracks over sea ice. Again, further investigation with labeled samples are necessary to explore these hypotheses.

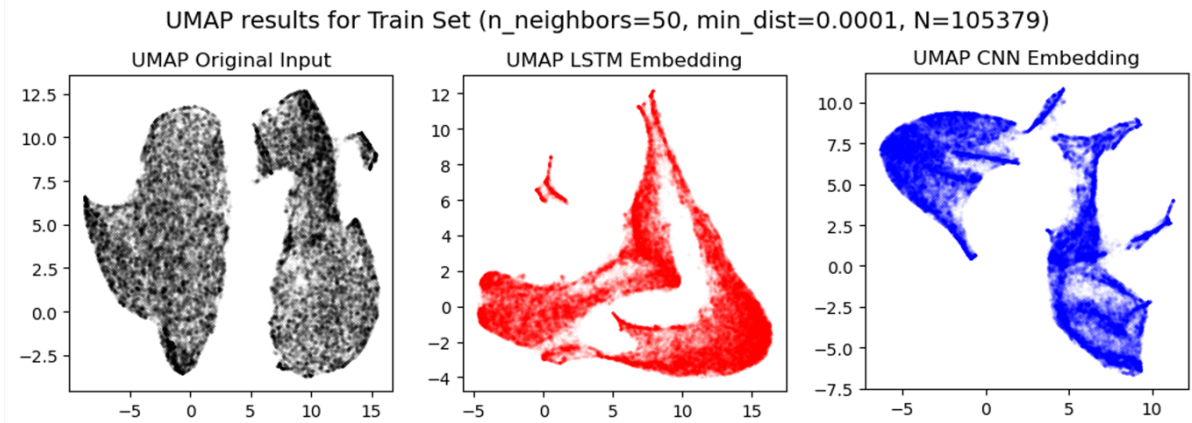

Fig. 4. UMAP visualization results Clustering results with Original Input of ATL07 (black) and embeddings from autoencoder models with LSTM (red) and CNN (blue).

Future studies will aim to assess the effectiveness of embeddings in downstream tasks, such as surface type classification. While UMAP was utilized for visualization in this study, a deeper comparison with clustering performed on the original embeddings without UMAP also necessary, given that the embeddings themselves perform dimensionality reduction. Investigating different seasons will provide critical insights into the effect of embeddings across seasonal variations, particularly during melting and freezing periods. Additionally, examining various sequence lengths and feature sets could help determine the optimal surface information required for better embeddings.

## IV. CONCLUSTION

This study explored the impact of learning embedding ICESat-2 data through unsupervised learning and clustering to assess its potential in addressing the limitation of small sample sizes. The autoencoder models using LSTM and CNN demonstrated high performance in reconstructing 100 m ATL07 sequences of height, photon rate, normalized background rate, and number of pulses. Compared to the original ATL07 input, the embeddings produced more compact clusters in the two-dimensional UMAP space. Our results suggest that the trained embeddings using unsupervised learning may capture meaningful information that may reduce the number of required samples for training, potentially enhancing clustering tasks. However, further investigation with labeled datasets is necessary to evaluate the practical effectiveness of ICESat-2 embeddings generated by the proposed approach.

ACKNOWLEDGMENT

This work was supported by the U.S. National Science Foundation award 2026962 and NASA award 80NSSC23K0936.

REFERENCES

[1] J. Stroeve and D. Notz, "Changing state of Arctic sea ice across all seasons," Environ. Res. Lett., vol. 13, no. 10, p. 103001, Oct. 2018.

[2] J. A. Screen and I. Simmonds, "The central role of diminishing sea ice in recent Arctic temperature amplification," Nature, vol. 464, no. 7293, pp. 1334–1337, Mar. 2010.

[3] T. Markus, et al., "The Ice, Cloud, and land Elevation Satellite-2 (ICESat-2): Science requirements, concept, and implementation," Remote Sens. Environ., vol. 190, pp. 260–273, Mar. 2017.

[4] R. Kwok, et al., "Surface height and sea ice freeboard of the Arctic Ocean from ICESat-2: Characteristics and early results," J. Geophys. Res. Oceans, vol. 124, no. 10, pp. 6942–6959, Oct. 2019.

[5] Y. Koo, et al., "Sea ice surface type classification of ICESat-2 ATL07 data by using data-driven machine learning model: Ross Sea, Antarctic as an example," Remote Sens. Environ., vol. 296, p. 113726, Mar. 2023.

[6] T. Chen, et al., "A simple framework for contrastive learning of visual representations," in Proc. Int. Conf. Mach. Learn. (ICML), Vienna, Austria, 2020, pp. 1597–1607.

[7] K. He, R. Girshick, and P. Dollár, "Rethinking ImageNet pre-training," in Proc. IEEE/CVF Int. Conf. Comput. Vis. (ICCV), Seoul, Korea, 2019, pp. 4918–4927

[8] E. P. Nowottnick, P. A. Selmer, and J. E. Yorks, "Autoencoders for denoising atmospheric profiles from ICESat-2," in Proc. 22nd Coherent Laser Radar Conf. (CLRC) / 31st Int. Laser Radar Conf. (ILRC), Landshut, Germany, Jun. 2024.

[9] W. Liu, M. Tsamados, A. Petty, T. Jin, W. Chen, and J. Stroeve, "Enhanced sea ice classification for ICESat-2 using combined unsupervised and supervised machine learning," Remote Sens. Environ., vol. 318, p. 114607, 2025.

[10] L. McInnes, J. Healy, and J. Melville, "UMAP: Uniform manifold approximation and projection for dimension reduction," arXiv preprint arXiv:1802.03426, 2018. [Online]. Available: https://arxiv.org/abs/1802.03426

[11] D. P. Kingma and J. Ba, "Adam: A method for stochastic optimization," arXiv preprint arXiv:1412.6980, 2014. [Online]. Available: https://arxiv.org/abs/1412.6980